\documentclass[runningheads]{llncs}
\usepackage{hyperref}
\usepackage{graphicx}
\usepackage{comment}
\usepackage{amsmath,amssymb} 
\usepackage{color}
\usepackage{xcolor}
\usepackage{wrapfig}
\usepackage{rotating} 
\usepackage{xcolor,pifont}
\newcommand*\colourcheck[1]{%
  \expandafter\newcommand\csname #1check\endcsname{\textcolor{#1}{\ding{52}}}%
}
\newcommand*\colourcross[1]{%
  \expandafter\newcommand\csname #1cross\endcsname{\textcolor{#1}{\ding{55}}}%
}
\colourcheck{blue}
\colourcheck{green}
\colourcheck{red}
\colourcross{red}

\usepackage{array}

\usepackage{lipsum}
\usepackage{blindtext}

\usepackage[font=small,labelfont=bf,tableposition=top]{caption}

\usepackage{algorithm,algpseudocode}
\usepackage{booktabs}
\usepackage{bbm}
\usepackage{multirow}
\usepackage{xcolor}
\usepackage[absolute]{textpos}

\begin{document}

\definecolor{somegray}{rgb}{0.5, 0.5, 0.5}
\newcommand{\darkgrayed}[1]{\textcolor{somegray}{#1}}
\begin{textblock}{8}(4, 0.7)
\begin{center}
\darkgrayed{This paper has been accepted for publication at the \\
European Conference on Computer Vision (ECCV), Glasgow, 2020}
\end{center}
\end{textblock}

\pagestyle{headings}
\mainmatter
\def\ECCVSubNumber{600}  

\title{Event-based Asynchronous \\ Sparse Convolutional Networks}


\titlerunning{Event-based Asynchronous Sparse Convolutional Networks}
%
\author{Nico Messikommer$^*$
\and Daniel Gehrig\thanks{Equal contribution}
\and Antonio Loquercio
\and \\ Davide Scaramuzza}
\authorrunning{N. Messikommer et al.}
%
\institute{Dept. Informatics, Univ. of Zurich and \\Dept. of Neuroinformatics, Univ. of Zurich and ETH Zurich}
{\let\newpage\relax\maketitle}
\begin{abstract}
Event cameras are bio-inspired sensors that respond to per-pixel brightness changes in the form of asynchronous and sparse ``events".
Recently, pattern recognition algorithms, such as learning-based methods, have made significant progress with event cameras by converting events into \emph{synchronous} dense, image-like representations and applying  traditional machine learning methods developed for standard cameras.
However, these approaches discard the spatial and temporal sparsity inherent in event data at the cost of higher computational complexity and latency. 
In this work, we present a general framework for converting models trained on synchronous image-like event representations into \emph{asynchronous} models with identical output, thus directly leveraging the intrinsic asynchronous 
and sparse nature of the event data.
We show both theoretically and experimentally that this drastically reduces the computational complexity and latency of high-capacity, synchronous neural networks
without sacrificing accuracy.
In addition, our framework has several desirable characteristics: (i) it exploits spatio-temporal sparsity of events explicitly, (ii) it is agnostic to the event representation, network architecture, and task, and (iii) it does not require any train-time change, since it is compatible with the standard neural networks' training process.
We thoroughly validate the proposed framework on two computer vision tasks: object detection and object recognition.
In these tasks, we reduce the computational complexity up to 20 times with respect to high-latency neural networks.
At the same time, we outperform state-of-the-art \emph{asynchronous} approaches up to $24\%$ in prediction accuracy.
\end{abstract}

\section*{Multimedia Material}
The code of this project is available at  \url{https://github.com/uzh-rpg/rpg_asynet}.
Additional qualitative results can be viewed in this video: \url{https://youtu.be/g_I5k_QFQJA} 
\section{Introduction}
Event cameras are \emph{asynchronous} sensors that operate radically differently from traditional cameras.
Instead of capturing dense brightness images at a fixed rate, event cameras measure brightness \emph{changes} (called \emph{events}) for each pixel independently.
Therefore, they sample light based on the scene dynamics, rather than on a clock with no relation to the viewed scene. 
By only measuring brightness changes, event cameras generate an asynchronous signal both sparse in space and time, usually encoding moving image edges~\footnote{https://youtu.be/LauQ6LWTkxM?t=4}~\cite{Gallego19arxiv}.
Consequently, they automatically discard redundant visual information and greatly reduce bandwidth.
In addition, event cameras possess appealing properties, such as
a very high dynamic range, high temporal resolution (in the order of 
microseconds), and low power consumption.

Due to the sparse and asynchronous nature of events, traditional computer vision algorithms cannot be applied, prompting the development of novel approaches. 
What remains a core challenge in developing these approaches is how to efficiently extract information from a stream of events. 
An ideal algorithm should maximize this information while exploiting the signal's spatio-temporal sparsity to allow for processing with minimal latency.

Existing works for processing event data have traded-off latency for prediction accuracy.
One class of approaches leverage filtering-based techniques to process events in sequence and thus provide predictions with high temporal resolution and low latency~\cite{Orchard15pami,Lagorce16pami,Gallego17pami,Kim16eccv}.
However, these techniques require significant engineering: event features and measurement update functions need to be handcrafted.
For this reason, they have difficulties in generalizing to many different tasks, especially high level ones as object recognition and detection.
Similarly, other works aim at reducing latency by making inference through a dynamical system, \emph{e.g.} a spiking neural network (SNN)\footnote{Here we use the term SNN as in the neuromorphic literature~\cite{Lee16fns}, where it describes continuous-time neural networks.
Other networks which are sometimes called SNNs are low precision networks, such as binary networks~\cite{Rastegari2016eccv}. However, these are not well suited for asynchronous inputs ~\cite{Lee16fns,PerezCarrasco13pami,Amir17cvpr}.}.
Despite having low latency, both filtering methods and SNNs achieve limited accuracy in high levels tasks, mainly due to their sensitivity to tuning and their difficult training procedure, respectively.
Recently, progress has been made by processing events in batches that are converted into intermediate input representations. 
Such representations have several advantages. Indeed, they have a regular, \emph{synchronous} tensor-like structure that makes them compatible with conventional machine learning techniques for image-based data (e.g. CNN). This has accelerated the development of new algorithms~\cite{Lagorce16pami,Sironi18cvpr,Maqueda18cvpr,Gehrig19iccv,Zhu18rss}.
In addition, it has been shown that many of these representations have statistics that overlap with those of natural images, enabling transfer learning with networks pretrained on image data~\cite{Maqueda18cvpr,Gehrig19iccv,Rebecq17ral,Rebecq19cvpr}.
This last class of approaches achieves remarkable results on several vision benchmarks but at the cost of discarding the asynchronous and sparse nature of event data.
By doing so they perform redundant computation at the cost of large inference times, thus losing the inherent low latency property of the event signal.
\subsubsection{Contributions}
We introduce a general event-based processing framework that combines the advantages of low latency methods and high accuracy batch-wise approaches.
Specifically, we allow a neural network to exploit the asynchronous and sparse nature of the input stream and associated representation, thus drastically reducing computation.
We mathematically show that the resulting asynchronous network generates identical results to its synchronous variant, while performing strictly less computation.
This gives our framework several desirable characteristics: (i) it is agnostic to the event representation, neural network architecture, and task; (ii) it does not require any change in the optimization or training process; (iii) it explicitly models the spatial and temporal sparsity in the events.
In addition, we demonstrate both theoretically and experimentally that our approach fully exploits the spatio-temporal sparsity of the data.
In order to do so, we relate our framework's computational complexity to the intrinsic dimensionality of the event signal, \emph{i.e.} the events' stream fractal dimension~\cite{xu2009viewpoint}.
To show the generality of our framework, we perform
experiments on two challenging computer vision tasks: object recognition
and object detection.
In these tasks, we match the performance of high capacity neural networks but with up 20 times less computation. 
However, our framework is not limited to these problems and can be applied without any change to a wide range of tasks.
\section{Related Work}
The recent success of data-driven models in frame-based computer vision~\cite{He16cvpr,Redmon16cvpr,Arandjelovic16cvpr} has motivated the event-based vision community to adopt similar pattern recognition models.
Indeed, traditional techniques based on handcrafted filtering methods~\cite{Orchard15pami,Lagorce16pami,Gallego17pami,Kim16eccv} have been gradually replaced with data-driven approaches using deep neural networks~\cite{Sironi18cvpr,Maqueda18cvpr,Gehrig19iccv,Zhu18rss,Rebecq19cvpr}.
However, due to their sparse and asynchronous nature, traditional deep models cannot be readily applied to event data, and this has sparked the development of several approaches to event-based learning. 
In one class of approaches, novel network architecture models directly tailored to the sparse and asynchronous nature of event-based data have been proposed \cite{Orchard15pami,Lee16fns,PerezCarrasco13pami,Amir17cvpr,Orchard13biocas,Lagorce17pami,Neil16nips}. 
These include spiking neural networks (SNNs) \cite{Orchard15pami,Lee16fns,PerezCarrasco13pami,Amir17cvpr,Orchard13biocas} which perform inference through a dynamical system by processing events as asynchronous spike trains.
However, due to their novel design and sensitivity to tuning, SNNs are difficult to train and currently achieve limited accuracy on high level tasks.
To circumvent this challenge, a second class of approaches has aimed at converting groups of events into image-like representations, which can be either hand-crafted~\cite{Lagorce16pami,Sironi18cvpr,Maqueda18cvpr,Zhu18rss} or learned with the task~\cite{Gehrig19iccv}. 
This makes the sparse and asynchronous event data compatible with conventional machine learning techniques for image data, \emph{e.g.} CNNs, 
which can be trained efficiently using back-propagation techniques.
Due to the higher signal-to-noise ratio of such representations with respect to raw event data, and the high capacity of deep neural networks, these methods achieve state-of-the-art results on several low and high level vision tasks \cite{Maqueda18cvpr,Gehrig19iccv,Zhu18rss,Zhu19cvpr,Rebecq19pami}.
However, the high performance of these approaches comes at the cost of discarding the sparse and asynchronous property of event data and redunant computation, leading to higher latency and bandwidth.
Recently, a solution was proposed that avoids this redundant computation by exploiting sparsity in input data~\cite{Graham18cvpr}.
Graham et al. proposed a technique to process spatially-sparse data efficiently, and used them to develop spatially-sparse convolutional networks.
Such an approach brings significant computational advantages to sparse data, in particular when implemented on specific neural network accelerators~\cite{Aimar19Nullhop}.
Sekikawa et al.~\cite{Sekikawa18costvel} showed similar computation gains when generalizing sparse operations to 3D convolutional networks.
However, while these methods can address the spatial sparsity in event data, they operate on synchronous data and can therefore not exploit the temporal sparsity of events.
This means that they must perform separate predictions for each new event, thereby processing the full representation at each time step.
For this reason previous work has focused on finding efficient processing schemes for operations in neural networks to leverage the temporal sparsity of event data.
Scheerlinck et al.~\cite{Scheerlinck19ral} designed a method to tailor the application of a single convolutional kernel, an essential building block of CNNs, to asynchronous event data.
Other work has focused on converting trained models into asynchronous networks by formulating efficient, recursive update rules for newly incoming events~\cite{Sekikawa19cvpr,Cannici19cvprw} or converting traditional neural networks into SNNs~\cite{Rueckauer17fns}.
However, some of these conversion techniques are limited in the types of representations that can be processed \cite{Cannici19cvprw,Rueckauer17fns} or lead to decreases in performance \cite{Rueckauer17fns}.
Other techniques rely on models that do not learn hierarchical features~\cite{Sekikawa19cvpr} limiting their performance on more complex tasks.
\newcommand{\kidx}{\mathbf{k}}
\newcommand{\kset}{\mathcal{K}}

\newcommand{\usite}{\mathbf{u}}
\newcommand{\aset}{\mathcal{A}}
\newcommand{\asite}{\mathbf{a}}
\newcommand{\isite}{\mathbf{i}}
\newcommand{\jsite}{\mathbf{j}}
\newcommand{\iset}{\mathcal{I}}
\newcommand{\cset}{\mathcal{C}}
\newcommand{\fset}{\mathcal{F}}
\section{Method}
In this section we show how to exploit the spatio-temporal sparsity of event data in classical convolutional architectures.
In Sec. \ref{sec:method:event_data} we introduce the working principle of an event camera.
Then, in Sec. \ref{sec:method:exploit_sparsity} we show how sparse convolutional techniques, such as Submanifold Sparse Convolutional (SSC) Networks \cite{Graham18cvpr}, can leverage this spatial sparsity.
We then propose a novel technique for converting standard \emph{synchronous networks}, into \emph{asynchronous networks} which process events asynchronously and with low computation.
\subsection{Event Data}
\label{sec:method:event_data}
Event cameras have independent pixels that respond to changes in the logarithmic 
brightness signal $L(\usite_k,t_k)\doteq\log I(\usite_k,t_k)$.
An event is triggered at pixel $\usite_k=(x_k,y_k)^T$ and at time $t_k$ as soon 
as the brightness increment since the last event at the pixel reaches a threshold 
$\pm C$ (with $C>0$):
\begin{equation}
    \label{eq:method:generative_model}
    L(\usite_k, t_k)-L(\usite_k, t_k-\Delta t_k) \geq p_k C
\end{equation}
where $p_k\in\{-1,1\}$ is the sign of the brightness change and $\Delta t_k$ is 
the time since the last event at $\usite$.
Eq.~\eqref{eq:method:generative_model} is the event generation model for an ideal sensor~\cite{Gallego17pami,Gallego15arxiv}.
During a time interval $\Delta \tau$ an event camera produces a sequence 
of events, $\mathcal{E}(t_N)=\{e_k\}_{k=1}^N=\{(x_k,y_k,t_k,p_k)\}_{k=1}^N$ with microsecond resolution.
Inspired by previous approaches \cite{Maqueda18cvpr,Gehrig19iccv,Zhu18rss,Tulyakov19iccv} we generate image-like representations $H_{t_N}(\usite, c)$  ($c$ denotes the channel) from these sequences, that can be processed by standard CNNs.
These representations retain the spatial sparsity of the events, since event cameras respond primarily to image edges, but discard their temporal sparsity.
Therefore, previous works only processed them synchronously, reprocessing them from scratch every time a new event is received.
This leads of course to redundant computation at the cost of latency.
In our framework, we seek to recover the temporal sparsity of the event stream by focusing on the change in $H_{t_{N}}(\usite, c)$ when a new event arrives:
\begin{equation}
    \label{eq:method:sparsely_updateable}
    H_{t_{N+1}}(\usite, c) = H_{t_{N}}(\usite, c) + \sum_{i}\Delta_i(c)\delta(\usite-\usite'_i). 
\end{equation}
This recursion can be formulated for arbitrary event representations, making our method compatible with general input representations.
However, to maximize efficiency, in this work we focus on a specific class of representations which we term \emph{sparse recursive representations} (SRR).
SRRs have the property that they can be \emph{sparsely updated} with each new event, leading to increments $\Delta_i(c)$ at only few positions $\usite'_i$ in $H_{t_{N}}(\usite, c)$. There are a number of representations which satisfy this criterion. 
In fact for the event histogram \cite{Maqueda18cvpr}, event queue \cite{Tulyakov19iccv}, and time image \cite{Mitrokhin18iros} only single pixels need to be updated for each new event.  
\begin{figure}[t]
    \centering
    \begin{tabular}{cccccc}
        \includegraphics[width=0.15\linewidth]{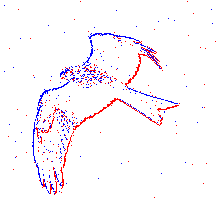}&
        \includegraphics[width=0.15\linewidth]{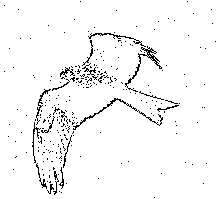}&
        \includegraphics[width=0.15\linewidth]{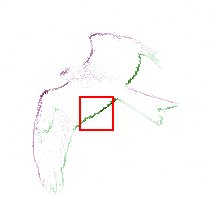}&
        \includegraphics[width=0.15\linewidth]{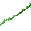} &
        \includegraphics[width=0.15\linewidth]{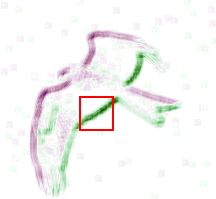} &
        \includegraphics[width=0.15\linewidth]{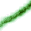}\\(a) events & (b) active sites&\multicolumn{2}{c}{(c) sparse activations}&\multicolumn{2}{c}{(d) dense activations}
    \end{tabular}
    \caption{Illustration of Submanifold Sparse Convolutions (SSC) \cite{Graham18cvpr}. A sparse event representation (a) is the input the network. SSCs work by only computing the convolution
    operation at active sites (b), \emph{i.e.} sites that are non-zero, leading to sparse activation maps in the subsequent layers (c). Regular convolutions on the other hand generate blurry 
    activation maps and therefore reduce sparsity (d).}
    \label{fig:method:sparse_vs_dense}
\end{figure}
\subsection{Exploiting the Sparsity of the Event Signal}
\label{sec:method:exploit_sparsity}
Event-cameras respond primarily to edges in the scene, which means that event representations are extremely sparse.
Submanifold Sparse Convolutions (SSC) \cite{Graham18cvpr}, illustrated in Fig. \ref{fig:method:sparse_vs_dense}, leverage spatial sparsity in the  data to drastically reduce computation.
Hence, they are not equivalent to regular convolutions.
Compared to regular convolutions, SSCs only compute the convolution at sites $\usite$ with a non-zero feature vector, and ignore inputs in the receptive field of the convolution which are 0. These sites with non-zero feature vector are termed \emph{active sites} $\aset_t$  (Fig. \ref{fig:method:sparse_vs_dense} (b)).
Fig.~\ref{fig:method:sparse_vs_dense} illustrates the result of applying an SSC to sparse event data (a). 
The resulting activation map (c) has the same active sites as its input and therefore, by induction, all SSC layers with the same spatial resolution share the same active sites and level of sparsity.
\begin{figure}[t]
    \centering
    \begin{tabular}{ccc}
        \includegraphics[width=0.3\linewidth]{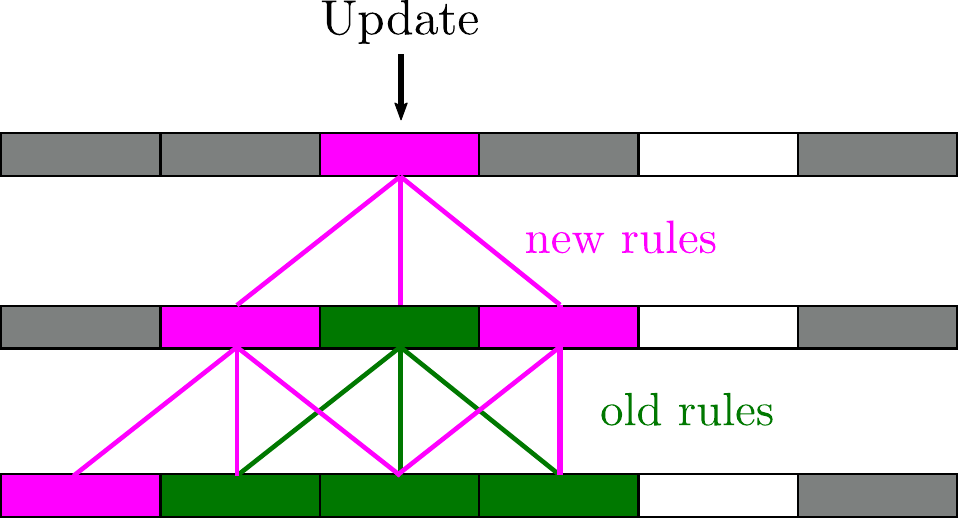}&
        \includegraphics[width=0.3\linewidth]{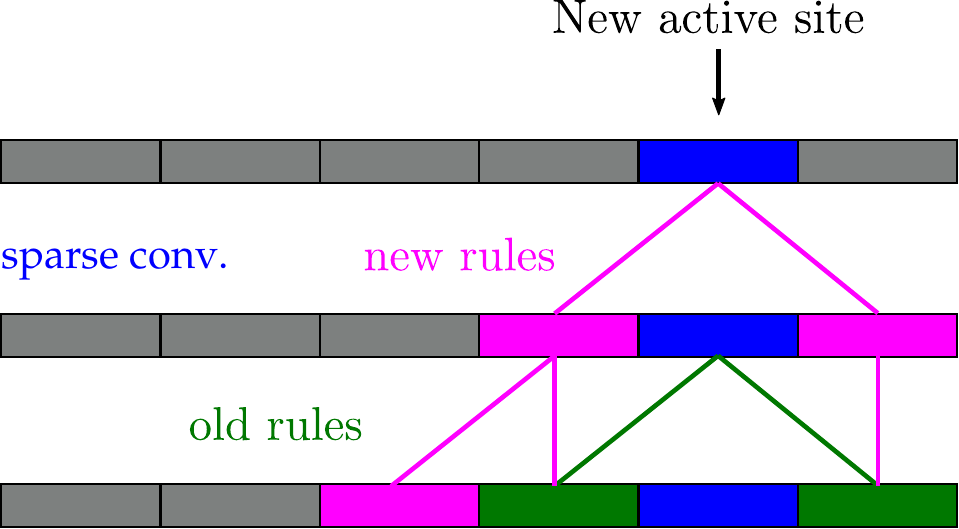}&
        \includegraphics[width=0.3\linewidth]{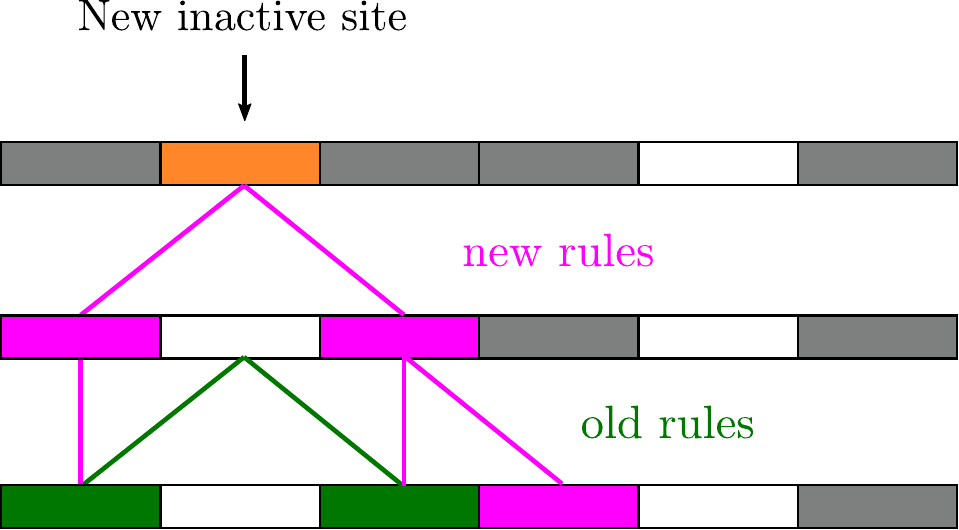}\\
        \multicolumn{3}{c}{\includegraphics[width=0.3\linewidth]{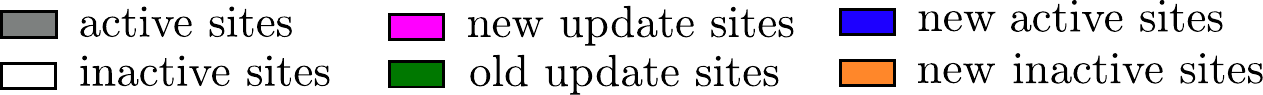}}\\
        (a) active site updated & (b) new active sites& (c) new inactive site
    \end{tabular}
    \caption{Propagation of the rulebook $\mathcal{R}_{\kidx,n}$ 
    a 1D example. The input is composed of active (gray) and inactive (white) sites. 
    (a) If the value of an active site changes (magenta), the update rules are incrementally added to the rulebook (lines) according to Eq. \eqref{eq:method:rulebook_prop}.
    (b) At newly active sites (blue) the sparse convolution is directly computed using Eq. \eqref{eq:method:single_layer_sparse_conv_active_sites} and repeated at each layer (here 1 to 3). (c) Similarly, new inactive sites (orange) are set to zero at each layer. Thus, new active sites (blue) and new inactive sites (orange) do not contribute to the rulebook propagation. 
    Best viewed in color.}
    \label{fig:method:rulebook}
\end{figure}
The sparse convolution operation can be written as 
\begin{align}
    \label{eq:method:single_layer_sparse_conv_active_sites}
    \tilde y_{n+1}^t(\usite,c)&=
        b_n(c)+\displaystyle\sum_{c'}\sum\limits_{\kidx\in\kset_n}\sum_{(\isite,\usite)\in R_{t,\kidx}}W_n(\kidx,c',c)y_n^t(\isite,c'), &\text{for } \usite\in\aset_t\\
    y_{n+1}^t(\usite,c)&=\sigma(\tilde y^t_{n+1}(\usite,c)).    
\end{align}
Here $y_n^t(\usite,c)$ is the activation of layer $n$ at time $t$ and is non-zero only for pixels $\usite$, $b_n$ denotes the bias term, $W_n$ and $\kidx\in\kset_n$ the parameters and indices of the convolution kernel, and $\sigma$ a non-linearity.
For the first layer $y_0^t(\usite, c)=H_{t_N}(\usite, c)$.
We also make use of the rulebook $R_{t,\kidx}$ \cite{Graham18cvpr}, 
a data structure which stores a list of correspondences $(\isite,\jsite)$ of input and output sites. 
In particular, a rule $(\isite, \jsite)$ is in $R_{t,\kidx}$ if both $\isite,\jsite\in\aset_t$ and $\isite-\jsite=\kidx$, meaning that the output $\jsite$ is in the receptive field of the input $\isite$ (Fig. \ref{fig:method:rulebook} (a) lines).
The activation at site $\isite$ is multiplied with the weight at index $\kidx$ and added to the activation at output site $\jsite$.
In \eqref{eq:method:single_layer_sparse_conv_active_sites}, this summation is performed over rules which have the same output site $\jsite=\usite$. 
When pooling operations such as max pooling or a strided convolution are encountered, the feature maps' spatial resolution changes and thus 
the rulebook needs to be recomputed. In this work we only consider max pooling. For sparse input it is the same as regular 
max pooling but over sites that are active. Importantly, after pooling, output sites become active when they have at least one active site in their receptive field. 
This operation increases the number of active sites.
\subsubsection{Asynchronous processing}
\label{sec:asyn_processing}
While SSC networks leverage the spatial sparsity of the events, they still process event representations synchronously, performing redundant computations.
Indeed, for each new event, all layer activations need to be recomputed. 
Since events are triggered at individual pixels, activations should also be affected locally. 
We propose to take advantage of this fact by retaining the previous activations $\{y_n^t(\usite,c)\}_{n=0}^N$ of the network and formulating novel, efficient update rules $y_n^t(\usite,c)\rightarrow y_n^{t+1}(\usite,c)$ for each new event.
By employing SRRs each new event leads to sparse updates $\Delta_i(c)$ at locations $\usite_i'$ in the input layer (Eq.~\eqref{eq:method:sparsely_updateable}). 
We propagate these changes to deeper layers by incrementally building a \emph{rulebook} $\mathcal{R}_{\kidx,n}$ and \emph{receptive field} $\mathcal{F}_n$ for each layer, visualized in Fig.~\ref{fig:method:rulebook}.
The rulebook (lines) are lists of correspondences $(\isite, \jsite)$ where $\isite$ at the input is used to update the value at $\jsite$ in the output. The receptive field (colored sited) keeps track of the sites that have been updated by the change at the input.
For sites that become newly active or inactive (Fig. \ref{fig:method:rulebook} (b) and (c)) the active sites $\aset_t$ are updated accordingly. 
At initialization (input layer) the rulebook is empty and the receptive field only comprises the updated pixel locations, caused by new events, \emph{i.e.} 
$\mathcal{R}_{\kidx,0}=\emptyset$ and $\mathcal{F}_0=\{\usite_i'\}_i$.
Then, at each new layer $\mathcal{R}_{\kidx,n}$ and $\mathcal{F}_n$ are expanded:
\begin{align}
    \label{eq:method:rulebook_prop}
    \mathcal{F}_{n}&=\{\isite-\kidx\vert \isite\in\mathcal{F}_{n-1}\text{ and }\kidx\in\kset_{n-1} \text{ if } \isite-\kidx\in \aset_t\}\\
    \mathcal{R}_{\kidx,n}&=\{(\isite,\isite-\kidx) \vert\isite \in \mathcal{F}_{n-1} \text{ if } \isite-\kidx\in \aset_t\}.
\end{align} 
\begin{figure}[t]
    \centering
    \begin{tabular}{cccccc}
        \includegraphics[height=0.15\linewidth]{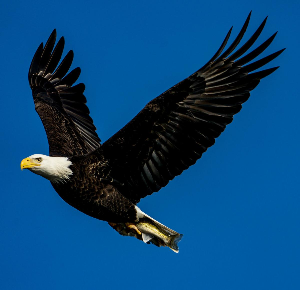}&
        \includegraphics[height=0.15\linewidth]{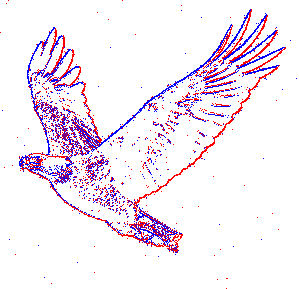}&
        \includegraphics[height=0.15\linewidth]{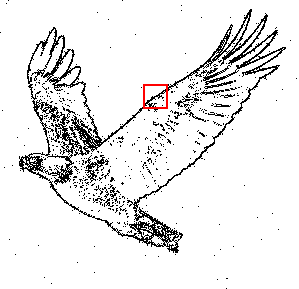}&
        \includegraphics[height=0.15\linewidth]{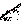} &
        \includegraphics[height=0.15\linewidth]{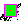} &
        \includegraphics[height=0.15\linewidth]{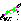}\\(a) image & (b) events & (c) active sites&(d) zoom &(e) dense&(f) sparse
    \end{tabular}
    \caption{The difference between asynchronous sparse and dense updates of events (b) is illustrated with an example image (a). 
    The active sites, \emph{i.e.} input sites that are non-zero, are visualized as black pixels (c). 
    (d)-(f) show an asynchronous update (red pixel) processed with traditional convolutions (e) and our proposed asynchronous sparse convolutions (f).
    The receptive field of traditional convolution (e) grows quadratically with network depth leading to redundant computation. 
    By contrast, our method (f) only updates active sites, which reduces the computational complexity. 
    In both cases the growth frontier is indicated in magenta.}
    \label{fig:method:receptive_field_growth}
\end{figure}
Rules that have a newly active or inactive site as output (Fig. \ref{fig:method:rulebook} (b) and (c), blue or orange sites) are ignored.\footnote{In the supplement we present an efficient recursive method for computing $\mathcal{R}_{\kidx,n}$ and $\mathcal{F}_n$ by reusing 
the rules and receptive field from the previous layers.}
We use Eq. \eqref{eq:method:rulebook_prop} to formulate the update rules to layer activations from time $t$ to $t+1$.
At the input we set $y_0^t(\usite,c)~=H_{t_N}(\usite,c)$ and then the update due to a single event can written as:
\begin{align}
    \label{eq:method:update_layer_n}
    \Delta_n(\usite, c) &= \sum_{\kidx\in\kset_{n-1}}\sum_{(\isite,\usite)\in \mathcal{R}_{\kidx,n}}\sum_{c'}W_{n-1}(\kidx,c',c)(y_{n-1}^{t}(\isite, c')-y_{n-1}^{t-1}(\isite, c'))\\
    \tilde y_{n}^{t}(\usite, c') &= \tilde y_{n}^{t}(\usite, c') + \Delta_n(\usite,c)\\
    y_{n}^{t}(\usite, c') &= \sigma(\tilde y_{n}^{t}(\usite, c')).
\end{align}
Note, that these equations only consider sites which have not become active (due to a new event) or inactive.
For newly active sites, we compute $y_n^t(\usite,c)$ according to Eq. \eqref{eq:method:single_layer_sparse_conv_active_sites}. 
Finally, sites that are deactivated are set to 0, \emph{i.e.} $y_n^t(\usite,c)=0$. In both cases we update the active sites $\aset_t$ before the update has been propagated.
By iterating over Eqs. \eqref{eq:method:update_layer_n} and \eqref{eq:method:rulebook_prop}, all subsequent layers can be updated recursively. 
Fig. \ref{fig:method:receptive_field_growth} illustrates the update rules above, applied to a single event update (red position) after six layers of both standard convolutions (e) and our approach (f).
Note that (e) is a special case of (f) with all pixels being active sites. 
By using our local update rules we see that computation is confined to a small patch of the image (c). 
Moreover, it is visible that standard convolutions (e) process noisy or empty regions (green and magenta positions), while our approach (f) focuses computation on sites with events, leading to higher efficiency.
Interestingly, we also observe that for traditional convolutions the size of the receptive field grows quadratically in depth while for our approach it grows more slowly, according on the \emph{fractal dimension} of the underlying event data. This point will be explored further in Sec. \ref{sec:method:computational_complexity}. 
\subsubsection{Equivalence of Asynchronous and Synchronous Operation}
By alternating between Eqs. \eqref{eq:method:rulebook_prop} and \eqref{eq:method:update_layer_n} asynchronous event-by-event updates can be propagated from the
input layer to arbitrary network's depth.
In the supplement we prove that processing $N$ events by this method is equivalent to processing all events at once and present pseudocode for our method.
It follows that a synchronous network, trained efficiently using back-propagation, can be deployed as an asynchronous network.
Therefore, our framework does not require any change to the optimization and learning procedure.
Indeed, any network architecture, after being trained, can be transformed in its asynchronous version, where it can leverage the high temporal resolution of the  events at limited computational complexity.
In the next section we explore this reduction in complexity in more detail.
\subsubsection{Computational Complexity}
\label{sec:method:computational_complexity}
\begin{figure}[t]
    \centering
    \begin{tabular}{cc}
        \includegraphics[height=0.40\linewidth]{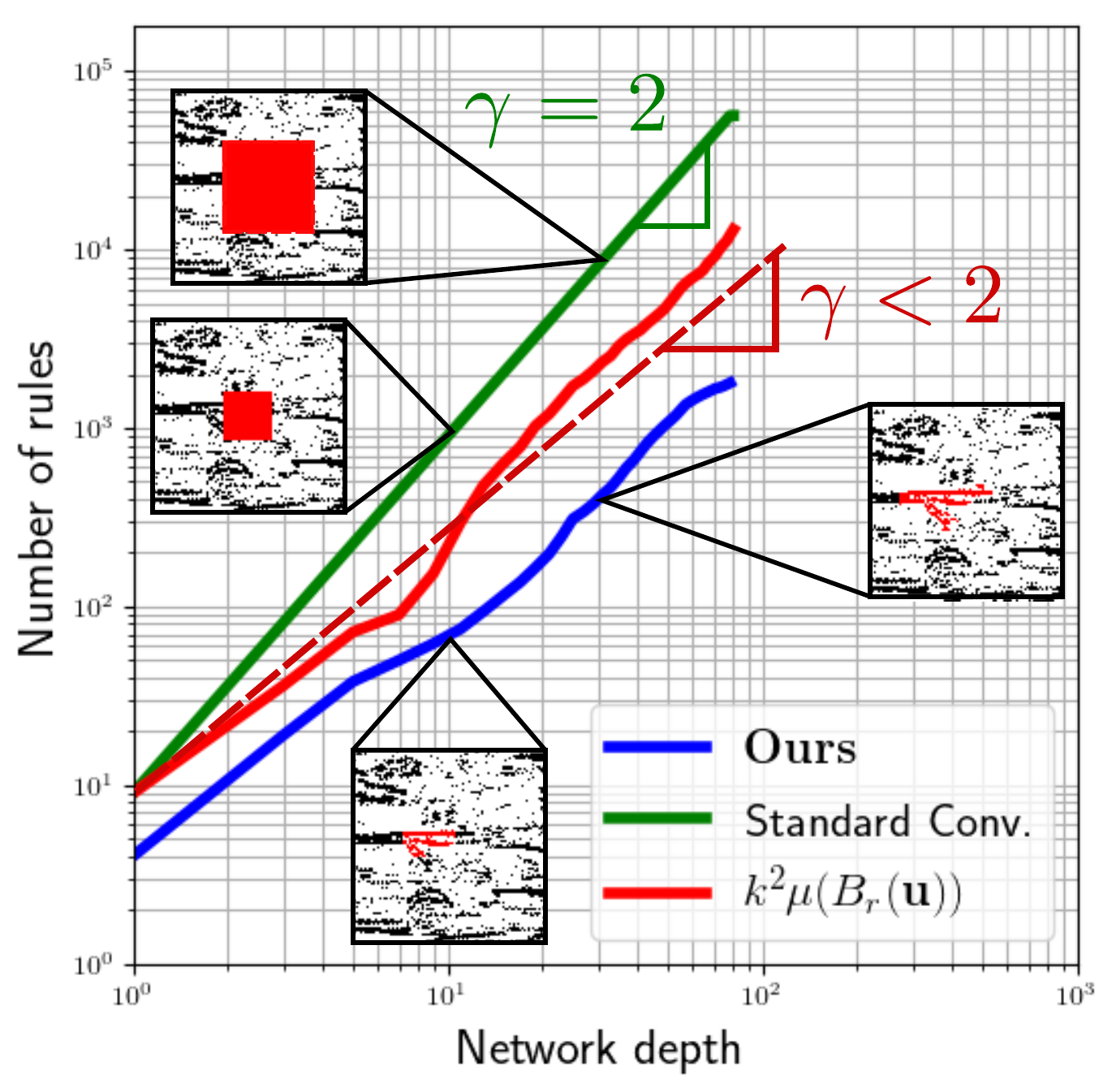}&
        \includegraphics[height=0.40\linewidth]{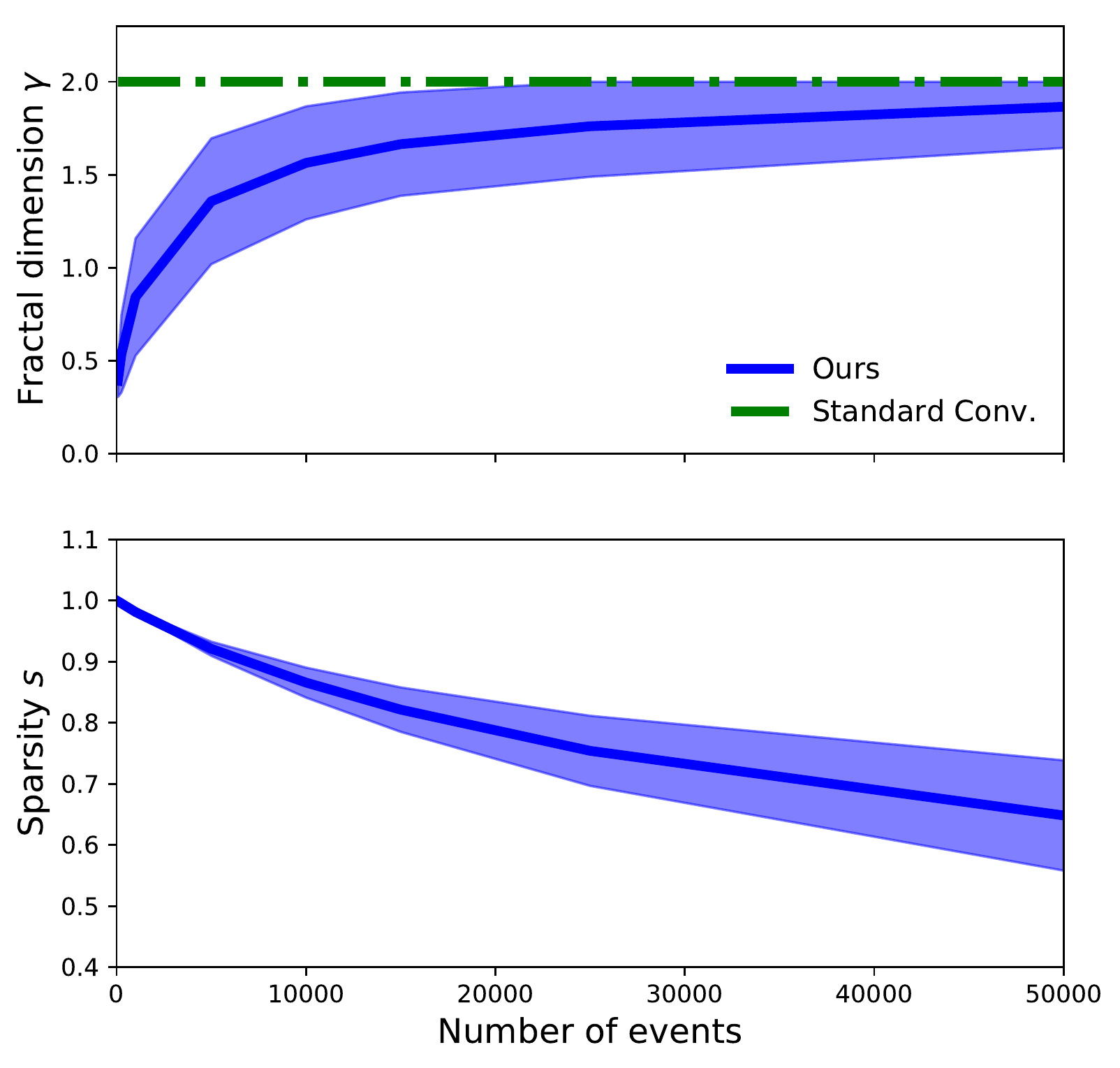}\\
        (a) & (b)
    \end{tabular}
    \caption{To bound the complexity of our method, we use the empirical sparsity $s$ and fractal dimension $\gamma$ of event data (b). \
    The fractal dimension characterizes the rate at which the number of update sites grows from one layer to the next. While for standard convolutions this number grows 
    quadratically (as $(1+(k-1)n)^2$, $k$ being the kernel size) with layer depth, with our method it grows more slowly, with an exponent $\gamma<2$.}
    \label{fig:method:fractal_dimension}
\end{figure}
In this section we analyze the computational complexity of our approach in terms of floating point operations (FLOPs), and compare it against conventional convolution operations.
In general, the number of FLOPs necessary to perform $L$ consecutive convolutions (disregarding non-linearities for the sake of simplicity) is:
\begin{equation}
    \label{eq:method:sparse_computation}
    C_{\text{dense}} =  \sum_{l=1}^L N(2k^2c_{l-1}-1)c_{l}
    \quad C_{\text{sparse}} =  \sum_{l=1}^L N^l_{r}(2c_{l}+1)c_{l-1}
\end{equation} 
These formulae are explained in more detail in the supplement.
For the sparse case, $N^l_{r}$ counts the number of rules, \emph{i.e.} input output pairs between layer $l-1$ and $l$, which corresponds exactly with the size of the rulebook in Eq. \eqref{eq:method:single_layer_sparse_conv_active_sites}.

Our method minimizes the number of rules it uses at each layer by only using a subset of the rules used by SSCs and incrementally expanding it from layer to layer (Fig. \ref{fig:method:rulebook}).
To characterize the computation from our approach we consider an update at a single pixel. 
At each layer the computation is proportional to the size of the rulebook in Eq. \eqref{eq:method:rulebook_prop} for which we can find an upper bound.
Let $n_l$ be the number of active pixels within a patch of size $1+(k-1)l$ which is an upper bound for the number of updated sites in layer $l$. If we assume that each pixel can have at most $k^2$ rules, the number of rules at layer $l$ is at most $n_lk^2$.

For a dense update, this number grows quadratically with the patch size $p=1+(k-1)l$, however, for sparse updates, this number grows more slowly.
To formalize this notion we define a measure $\mu(B(\usite,r)))$ which counts the number of active sites within a patch of radius $r=\frac{p}{2}$.
This measure can be used to define the \emph{fractal dimension} of event data at pixel $\usite$ according to \cite{Xu09ijcv}: 
\begin{equation}
    \gamma(\usite)=\lim_{r\rightarrow0}\frac{\log(\mu(B(\usite,r)))}{\log {2r}}
\end{equation}
The fractal dimension describes an intrinsic property of the event event data, related to its dimensionality and has not been characterized for event data prior to this work. 
It measures the growth-rate of the number of active sites as the patch size is varied. In particular, it implies that this number 
grows approximately as $n_l\approx (1+(k-1)l)^\gamma$.
To estimate the fractal dimension we consider the slope of $\mu(B(\usite,r)))$ over $r=\frac{1+(k-1)l}{2}$ in the log-log domain which we visualize in Fig.~\ref{fig:method:fractal_dimension}.
Crucially, a slope $\gamma<2$ indicates that the growth is slower than quadratic.
This highlights the fact that event data exists in a submanifold of the image plane ($\gamma=2$) which has a lower dimension than two.
With this new insight we can find an upper bound for the computation using Eq. \eqref{eq:method:sparse_computation}:
\begin{equation}
    \label{eq:method:async_sparse_computation}
    C_{\text{async. sparse}}\leq \sum_{l=1}^L c_{l-1} \left(2c_{l}+1\right)n_lk^2 \approx \sum_{l=1}^L c_{l-1}\left(2c_{l}+1\right)(1+(k-1)l)^\gamma k^2
\end{equation}
Where we have substituted the rulebook size at each layer.
At each layer, our method performs at most $k^2c_{l-1} (2c_{l}+1)\left(1+(k-1)l\right)^\gamma$ FLOPs. If we compare this with the per layer computation used by a dense network (Eq. \eqref{eq:method:sparse_computation}) we see that our method performs significantly less computation:
\begin{equation}
    \frac{C^l_{\text{sparse. anync}}}{C^l_{\text{dense}}} \leq \frac{k^2(2c_{l}+1)c_{l-1}}{(2k^2c_{l-1}-1)c_{l}}\frac{\left(1+(k-1)l\right)^\gamma}{N}\approx\frac{\left(1+(k-1)l\right)^\gamma}{N}<<1
\end{equation}
Where we assume that $2c_l>>1$ and $k^2c_{l-1}>>1$ which is the case in typical neural networks\footnote{In fact, for typical channel sizes $c_l\geq16$ we incur a $<3\%$ approximation error}.
Moreover, as the fractal dimension decreases our method becomes exponentially more efficient.
Through our novel asynchronous processing framework, the fractal dimension of event data can be exploited efficiently. 
It does so with sparse convolution, that can specifically process
low-dimensional input data embedded in the image plane, such as points and lines.

\section{Experiments}
We validate our framework on two computer vision applications: object recognition (Sec.~\ref{sec:obj_class}) and object detection (Sec.~\ref{sec:obj_det}).
On these tasks, we show that our framework achieves state-of-the-art results with a fraction of the computation of top-performing models.
In addition, we demonstrate that our approach can be applied to different event-based representations.
We select the event histogram~\cite{Maqueda18cvpr} and the event queue~\cite{Tulyakov19iccv} since they can be updated sparsely and asynchronously for each incoming event (see Sec.~\ref{sec:asyn_processing}).
\newcolumntype{C}[1]{>{\centering}m{#1}}
\subsection{Object Recognition}
\label{sec:obj_class}
We evaluate our method on two standard event camera datasets for object recognition: Neuromorphic-Caltech101 (N-Caltech101)~\cite{Orchard15fns}, and N-Cars~\cite{Sironi18cvpr}.
The N-Caltech101 dataset poses the task of event-based classification of 300 ms sequences of events. 
In total, N-Caltech101 contains 8,246 event samples, which are labelled into 101 classes.
N-Cars~\cite{Sironi18cvpr} is a benchmark for car recognition.
It contains 24,029 event sequences of 100 ms which contain a car or a random scene patch.
To evaluate the computational efficiency and task performance we consider two metrics: prediction accuracy and number of floating point operations (FLOPs).
While the first indicates the prediction quality, the second one measures the computational complexity of processing an input.
The number of FLOPs is commonly used as a complexity metric~\cite{He16cvpr,Graham18cvpr,Rueckauer17fns}, since it is independent of both the hardware and the implementation.
Details of the FLOP computation are reported in the supplement.
Analogously to previous work on sparse data processing~\cite{Graham18cvpr}, we use a VGG13~\cite{Simonyan2015iclr} architecture with 5 convolutional blocks and one final fully connected layer.
Each block contains two convolution layers, followed by batch-norm~\cite{Ioffe15icml}, and a max pooling layer. 
We train the networks with the cross-entropy loss and the ADAM optimizer~\cite{Kingma15iclr}.
The initial learning rate of $10^{-4}$ is divided by 10 after 500 epochs.
\begin{figure}[t]
    \centering
    \small
    \begin{tabular}{ccc}
        \includegraphics[height=0.25\linewidth]{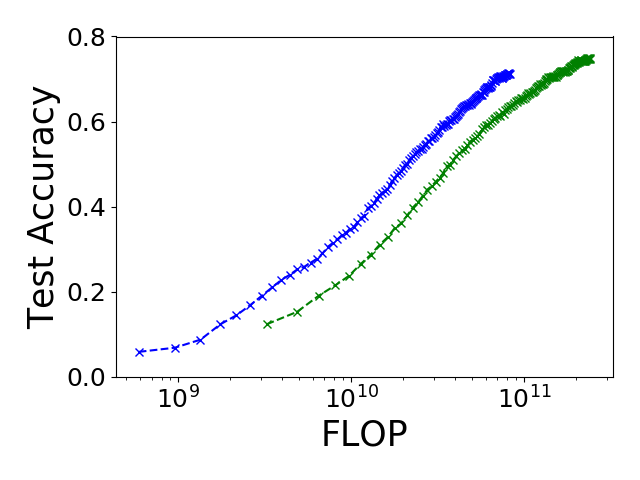}&
        \includegraphics[height=0.25\linewidth]{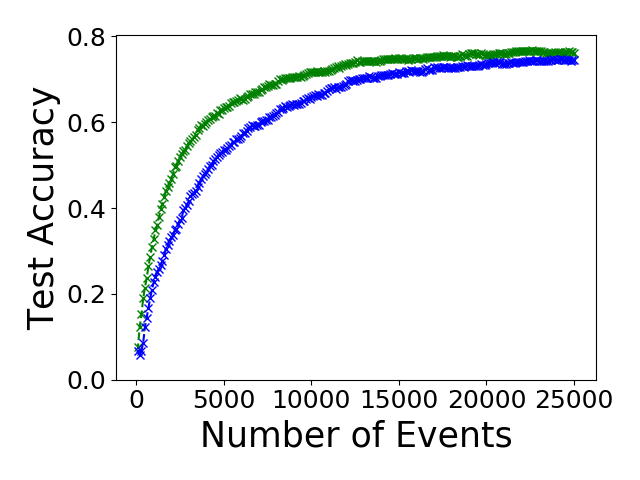}&
        \includegraphics[height=0.25\linewidth]{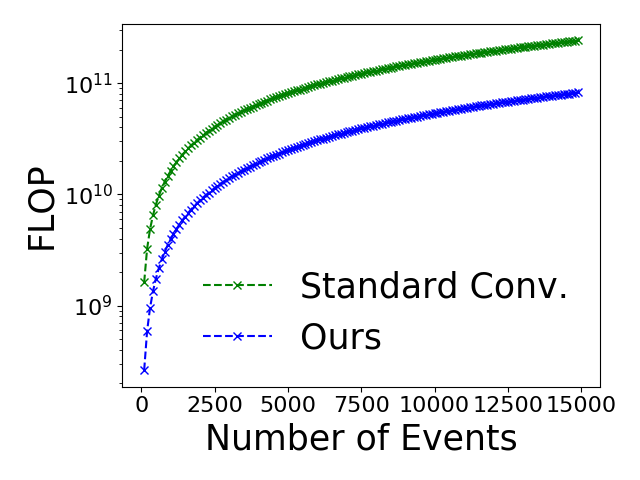}\\
        (a) & (b) & (c)
    \end{tabular}
    \caption{\small Our approach requires less cumulative FLOPs w.r.t. the dense method (Standard Conv.) to produce similarly accurate predictions (a). Although dense processing requires fewer events to generate predictions of equal accuracy (b), it needs significantly more computation per event, thus having higher cumulative FLOPs~(c).}
    \label{fig:experimental:flop_accuracy}
\end{figure}
\subsubsection{Results}
In our first experiment, we compare our sparse and asynchronous processing scheme to the dense and synchronous one.
For comparability, both methods share the same VGG13 architecture and the same input representation, which was generated with $25,000$ events. 
This number was empirically found to yield a good trade-off between accuracy and computational efficiency (see supplement).
We measure the approaches' computational complexity in terms of required FLOPs per single event update.
Classification results shown in Tab.~\ref{tab:obj_clas} demonstrate that our processing scheme has similar accuracy to the dense one but requires up to 19.5 times less computations per event.
\newcommand{\pre}[1]{{\color{blue}#1}}
\begin{table}
\centering
\resizebox{0.9\textwidth}{!}{
\renewcommand{\arraystretch}{1.1}
\begin{tabular}{m{2.5cm}C{3cm}C{2.0cm}C{2.0cm}C{2.0cm}>{\centering\arraybackslash}m{2.0cm}}
 &  &\multicolumn{2}{c}{N-Caltech101} & \multicolumn{2}{c}{N-Cars} \\
  \cmidrule(lr){3-4} \cmidrule(lr){5-6}
 & Representation &  Accuracy $\uparrow$ & MFLOP/ev $\downarrow$ & Accuracy $\uparrow$ & MFLOP/ev $\downarrow$\\
 \hline
 Standard Conv. & \multirow{2}{*}{Event Histogram} & \textbf{0.761}& 1621& \textbf{0.945}& 321\\
 Ours & & 0.745& \textbf{202} &0.944& \textbf{21.5}\\
 \hline
 Standard Conv. & \multirow{2}{*}{Event Queue} & 0.755& ~2014& 0.936& 419\\
 Ours & & 0.741& \textbf{202} & 0.936& \textbf{21.5}\\
\end{tabular}}
\caption{\label{tab:obj_clas} Our approach matches the performance of the traditional dense and synchronous processing scheme at one order of magnitude less computations.}
\end{table}
The low-latency of the event signal allows us to make fast predictions. For this reason we compare the maximal prediction accuracy that can be achieved given a fixed computation budget for our method and a standard CNN. 
This experiment imitates the scenario where an object suddenly appears in the field of view, \emph{e.g.} a pedestrian crossing the street.
To do this we use samples from N-Caltech101 \cite{Orchard15fns} and report the multi-class prediction accuracy (Fig. \ref{fig:experimental:flop_accuracy} (b)) and total number of FLOPs used (Fig. \ref{fig:experimental:flop_accuracy} (c)) as a function of number of events observed for our method and a standard CNN with the same architecture.
For each new event we compute the FLOPs needed to generate a new prediction and its accuracy, taking into account all previously seen events.
It can be seen that both standard CNN and our method have a higher prediction accuracy as the number of events increases (Fig. \ref{fig:experimental:flop_accuracy} (b)). 
However, compared to standard networks our method performs far less computation (Fig. \ref{fig:experimental:flop_accuracy} (c)). This is because for standard networks all activations need to be recomputed for each new event, while our method retains the state and therefore only needs to perform sparse updates. 
As a result, our method achieves up to $14.8\%$ better accuracy at the same FLOP budget (Fig.~\ref{fig:experimental:flop_accuracy} (a)), thus improving the prediction latency.
Moreover, to show the flexibility of our approach to different update rates, we initialize a representation with $25,000$ events and update it either for each new event or for a batch of 100 new events.
For comparability, all methods share the same VGG13 architecture and input representation.
\begin{figure*}[t]
     \centering
     \includegraphics[width=\linewidth]{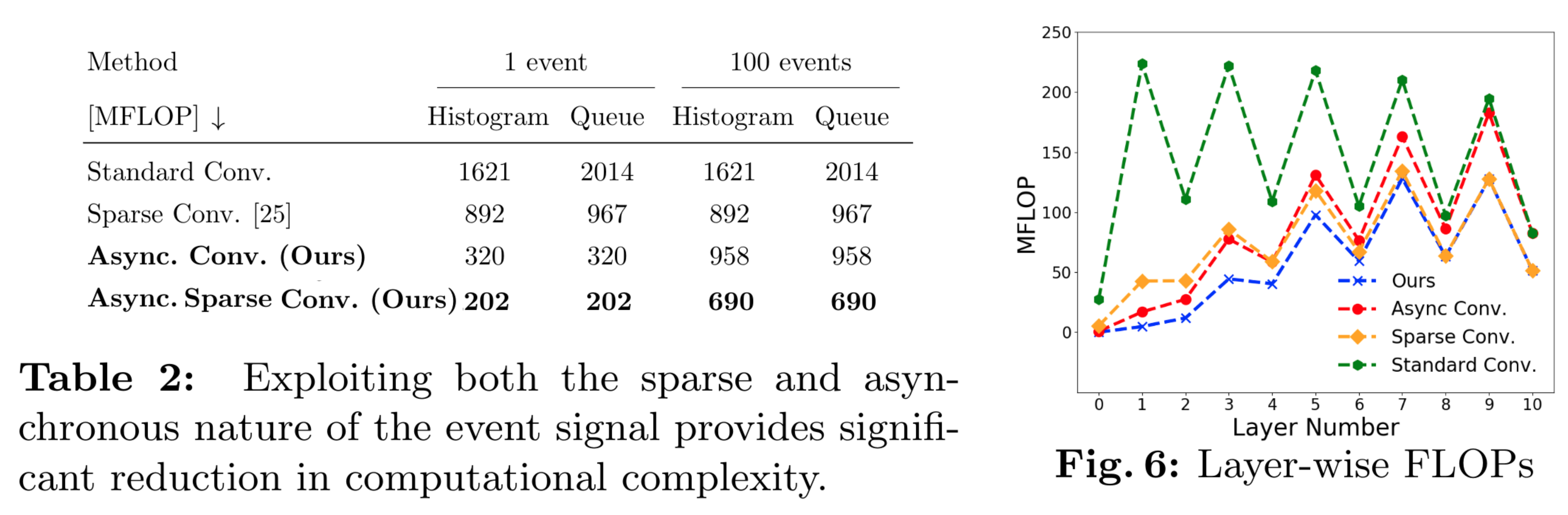}
   \end{figure*}
\stepcounter{figure}
\stepcounter{table}
The results of this experiment are reported in Tab. 2.
For both single and batch event update, exploiting the asynchronous and sparse nature of the signal offers significant computational benefits.
Our approach (\emph{Async. Sparse Conv.}) performs on average 8.9 times less computation than standard convolution for one event update, and of 2.60 times for 100 events update.
We additionally evaluate the computation per layer of each of previous models in the case of one event update.
Fig.~6, which presents the results of this evaluation, shows that sparse and asynchronous processing saves the majority of computations in the initial layers of the network.
Indeed, the input representations of the event stream are spatially very sparse (see Fig.~\ref{fig:method:fractal_dimension}(b)), and are only locally updated for each new event.
\begin{table}
\small
\centering
\resizebox{0.8\textwidth}{!}{
\renewcommand{\arraystretch}{1.0}
\begin{tabular}{m{1.6cm}C{1.5cm}C{1.8cm}C{2.1cm}C{1.8cm}>{\centering\arraybackslash}m{2.1cm}}
 & &\multicolumn{2}{c}{N-Caltech101} & \multicolumn{2}{c}{N-Cars} \\
  \cmidrule(lr){3-4} \cmidrule(lr){5-6}
   Methods& Async.& Accuracy $\uparrow$ & MFLOP/ev $\downarrow$ & Accuracy $\uparrow$ & MFLOP/ev $\downarrow$ \\
 \hline
 H-First \cite{Orchard15pami}   & \greencheck&0.054 & - & 0.561 & -\\
 HOTS \cite{Lagorce17pami}     & \greencheck& 0.210 & 54.0 & 0.624 & 14.0\\
 HATS  \cite{Sironi18cvpr}    & \greencheck& 0.642 & \textbf{4.3} & 0.902 & \textbf{0.03}\\
 DART \cite{Ramesh17arxiv}     & \greencheck& 0.664 & - & - & -\\
 YOLE \cite{Cannici19cvprw}    & \greencheck& 0.702 & 3659 & 0.927 & 328.16 \\
 EST \cite{Gehrig19iccv}    &\redcross& \textbf{0.817} & 4150 & 0.925 & 1050 \\
  SSC \cite{Graham18cvpr} &\redcross& 0.761& 1621& \textbf{0.945}& 321\\
    \hline
  \textbf{Ours} & \greencheck& \textbf{0.745 }& 202 &\textbf{
 0.944} & 21.5\\
\end{tabular}}
\caption{\label{tab:class_sota} Comparison with  asynchronous and dense methods for object recognition.}
\end{table}
\subsubsection{Comparison with State-of-the-Art Methods} 
We finally compare our approach with state-of-the-art methods for event-based object recognition.
We consider models that, like ours, perform efficient per-event updates due to a light-weight computational model (HATS~\cite{Sironi18cvpr}, HOTS~\cite{Lagorce17pami}, DART~\cite{Ramesh17arxiv}), a spiking neural network (H-First~\cite{Orchard15pami}), or asynchronous processing (YOLE~\cite{Cannici19cvprw}).
The results for this evaluation are presented in Tab.~\ref{tab:class_sota}.
Our method outperforms the state-of-the-art (YOLE) by 4.3\% in accuracy on N-Caltech101 and 1.7\% on N-Cars at only 6\% (on average over datasets) of its computational cost.
Finally, we compare against the synchronous state-of-the-art method \cite{Gehrig19iccv}. Our method achieves a slightly higher accuracy on N-Cars using only 21.5 MFLOPs (47 times reduction). Similarly, our method using the asynchronous processing requires 20 times fewer FLOPs on N-Caltech101 but at the cost of lower accuracy.
In addition to the performance evaluation, we timed our experiments conducted on N-Caltech101 by measuring the processing time for a single event on a i7-6900K CPU. 
In our framework implemented in C++ and Python, our method requires 80.4 ms, while the standard dense CNN needs 202 ms.
Therefore, our approach becomes roughly 2.75 times faster by leveraging the sparsity. However, in the highly-optimized framework PyTorch \cite{Paszke17nipsw}, a dense inference takes only 23.4 ms. 
Given its lower number of FLOPs, we expect that our method will experience a significant run-time reduction with further optimizations and specific hardware.
\subsection{Object Detection}
\label{sec:obj_det}
Object detection is the task of regressing a bounding box and class probabilities for each object in the image. 
We evaluate our method on two standard benchmarks for event-based object detection: N-Caltech101~\cite{Orchard15fns} and Gen1 Automotive~\cite{tournemire2020large}.
While the former contains the N-Caltech101 samples each with a single bounding box, the latter contains 228,123 bounding boxes for cars and 27,658 for pedestrians collected in an automotive scenario.
For this task we combine the first convolutional blocks of the object recognition task with the YOLO output layer~\cite{redmon2015look}.
The resulting feature maps are processed by the YOLO output layer to generate bounding-boxes and class predictions.
As is standard, we report the performance using the mean average precision (mAP) metric~\cite{Ever10ijcv} using the implementation of~\cite{Padilla20IWSSIP}.
As for the recognition task, we measure computational complexity with FLOPs.
\begin{table}
\centering
\resizebox{0.9\textwidth}{!}{
\renewcommand{\arraystretch}{1.1}
\begin{tabular}{m{2.5cm}C{3cm}C{1.5cm}C{1.5cm}C{1.5cm}>{\centering\arraybackslash}m{1.5cm}}
 &  &\multicolumn{2}{c}{N-Caltech101} & \multicolumn{2}{c}{Gen1 Automotive} \\
  \cmidrule(lr){3-4} \cmidrule(lr){5-6}
 & Representation &  mAP $\uparrow$ & MFLOP $\downarrow$ & mAP $\uparrow$ & MFLOP $\downarrow$\\
 \hline
  YOLE \cite{Cannici19cvprw} & Leaky Surface & 0.398 & 3682& -& -\\
  \hline
Standard Conv. & \multirow{2}{*}{Event Queue} & 0.619& 1977& \textbf{0.149}& 2614\\
Ours & & 0.615& \textbf{200}& 0.119&  \textbf{205}\\
 \hline
Standard Conv. & \multirow{2}{*}{Event Histogram} & 0.623 & 1584& 0.145& 2098\\
 Ours & & \textbf{0.643}& \textbf{200}&  0.129&  \textbf{205}\\
\end{tabular}}
\caption{\label{tab:obj_det} Accuracies for object detection.}
\end{table}
\subsubsection{Results and Comparison with State-of-the-Art}
Tab.~\ref{tab:obj_det} shows quantitative results on object detection while Fig. 1 in the supplement illustrates qualitative results.
Our approach achieves comparable or superior performance with respect to standard networks at a fraction of the computational cost.
Specifically, for the histogram representation our method outperforms dense methods by 2.0\% on N-Caltech101. On the Gen1 Automotive dataset, we experience a slight performance drop of 1.8\%.
The slight performance improvement in N-Caltech101 is probably thanks to the sparse convolution, which give less weight to noisy events.
In terms of computation, our method reduces the number of FLOPs per event by a factor of 10.6 with respect to the dense approach, averaged over all datasets and representations.
Tab.~\ref{tab:obj_det} also compares our approach to the state-of-the-art method for event-based detection, YOLE~\cite{Cannici19cvprw}.
Compared to this baseline, we achieve 24.5\% higher accuracy at 5\% of the computational costs.

\section{Discussion}
In the quest of high prediction accuracy, event-based vision algorithms have relied heavily on processing events in synchronous batches using deep neural networks.
However, this trend has disregarded the sparse and asynchronous nature of event-data.
Our work brings the genuine spatio-temporal sparsity of events back into high-performance CNNs, by significantly decreasing their computational complexity (up to 20 times).
By doing this, we achieve up to 15\% better accuracy than state-of-the-art synchronous models at the same computational (FLOP) budget.
In addition, we outperform existing asynchronous approaches by up to 24.5\% in accuracy.
Our work highlights the importance for researchers to take into account the intrinsic properties of event data in the pursuit of low-latency and high-accuracy event vision algorithms.
Such considerations will open the door to unlock the advantages of event
cameras on all tasks that rely on low-latency visual information processing.
\section{Acknowledgements}
This  work  was  supported  by  the  Swiss  National  Center of Competence Research Robotics (NCCR), through the Swiss  National  Science  Foundation,  and  the  SNSF-ERC starting grant.
\bibliographystyle{IEEEtran}
\bibliography{rpg_bib/all}
\clearpage
\section{Supplementary Material}
In the supplementary material, references which point to the main manuscript will be referenced with a leading "M-".  
In Sec. \ref{sec:app:efficient_rulebook_update} we describe an efficient recursive method for computing the rulebook $\mathcal{R}_{\kidx,n}$ in 
Eq. (M-7) and present the asynchronous propagation of changes through events in algorithmic form in Tab. \ref{alg:app:sparse_con1}. In Sec. \ref{sec:app:proof_equivalence} we present a proof of the equivalence of network outputs using asynchronous and synchronous networks. In Sec. \ref{sec:app:representations} we present additional details about the input representations and FLOP calculations used in the experiments in Sec. M-4.
In Sec. \ref{sec:app:ablation} we present a sensitivity analysis where we vary the number of events used for training and justify our choice of 25'000 events for all experiments. Finally, in Sec. \ref{sec:qual_results_object_detection} we show additional qualitative object detection results.
\subsection{Efficient Rulebook Update}
\label{sec:app:efficient_rulebook_update}
At each layer the rulebook $\mathcal{R}_{\kidx,n}$ and receptive field $\mathcal{F}_n$ are
\begin{align}
    \nonumber\mathcal{F}_{n}&=\{\isite-\kidx\vert \isite\in\mathcal{F}_{n-1}\text{ and }\kidx\in\kset_{n-1} \text{ if } \isite-\kidx\in \aset_t\}\\
    \nonumber\mathcal{R}_{\kidx,n}&=\{(\isite,\isite-\kidx) \vert\isite \in \mathcal{F}_{n-1} \text{ if } \isite-\kidx\in \aset_t\}.
\end{align} 
At the input these are initialized as $R_{\kidx,0}=\emptyset$ and $\mathcal{F}_0=\{\usite'_i\}$.
The propagation of these two data structures is illustrated in Fig. M-2. 
We observe that at each layer the rulebook and receptive field can be computed by reusing the data from the previous layer. 
We can do this by decomposing the receptive field into a frontier set $f_n$ (Fig. M-2 (a) magenta sites)and visited state set $F_n$ (Fig. M-2 (a) green sites). At each layer $\mathcal{F}_n=f_n\cup F_n$.
To efficiently update both $\fset_n$ and $\mathcal{R}_{\kidx,n}$ at each layer we only consider the rules that are added due to inputs in the frontier set (Fig. M-2 (a), magenta lines). These can be appended to the existing rulebook. In addition, the receptive field $\fset_n$ can be updated similarly, by adding new update sites reached from the frontier set. This greatly reduces the sites that need to be considered in the computation of $\mathcal{R}_{\kidx,n}$ and $\fset_n$ in Eqs. (M-6) and (M-7).   

\subsection{Equivalence of Synchronous and Asynchronous Updates}
\label{sec:app:proof_equivalence}
We start with Eq. (M-4), which we repeat here:
\begin{align*}
    \tilde y_{n+1}^t(\usite,c)&=
    \begin{cases}
        b_n(c)+\displaystyle\sum_{c'}\sum\limits_{\kidx\in\kset_n}\sum_{(\isite,\usite)\in R_{t,\kidx}}W_n(\kidx,c',c)y_n^t(\isite,c'), &\text{for } \usite\in\aset_t\\
        0 &\text{ else }
    \end{cases}\\
    y_{n+1}^t&=\sigma(\tilde y^t_{n+1})    
\end{align*}
Here the input layer is $y_0^t(\usite,c)=H_{t_{N}}(\usite,c)$. In a next step we assume changes to the input layer as in Eq. (M-3). 
These changes occur at sites $\usite_i\in\fset_0$ with magnitude $\Delta_i(c)=y_0^{t+1}(\usite_i,c)-y_0^{t}(\usite_i,c)$. The sites $\usite_i$ can be categorized into three groups: sites that are and remain active, sites that become inactive (feature becomes 0) and sites that become active. We will now consider how $y_n^{t+1}(\usite,c)$ evolves:
\begin{align}
\label{eq:app:update}
    \tilde y_{1}^{t+1}(\usite,c)&= b_0(c)+\displaystyle\sum_{c'}\sum\limits_{\kidx\in\kset_0}\sum_{(\isite,\usite)\in R_{t+1,\kidx}}W_n(\kidx,c',c)y_0^{t+1}(\isite,c')\\
    &=b_0(c)+\displaystyle\sum_{c'}\sum\limits_{\kidx\in\kset_0}\sum_{(\isite,\usite)\in R_{t+1,\kidx}}W_0(\kidx,c',c)\left(y_0^{t}(\isite,c')+\Delta(\isite,c')\right)\\
\end{align}
Here we define the increment $\Delta_0(\usite, c)$.
This increment is only non-zero for sites at which the input $y_0^t(\isite,c)$ changed, so for $(\isite,\usite)\in R_{t+1,\kidx}$ such that $\isite\in\fset_0$. 
At time $t+1$ the rulebook $R_{t+1,\kidx}$ is modified for every site $\usite_j$ that becomes newly active:
\begin{align*}
    R_{t+1,\kidx}=R_{t,\kidx}\cup \bigcup_j \{(\usite_j+\kidx, \usite_j)\vert \usite_j+\kidx\in\aset_{t+1}\}\cup\{(\usite_j, \usite_j-\kidx)\vert \usite_j-\kidx\in\aset_{t}\}
\end{align*}
and every site $u_l$ that becomes inactive
\begin{align*}
    R_{t+1,\kidx}=R_{t,\kidx}\backslash\bigcup_l \{(\usite_l+\kidx, \usite_l)\vert \usite_l+\kidx\in\aset_{t}\}\cup\{(\usite_l, \usite_l-\kidx)\vert \usite_l-\kidx\in\aset_{t+1}\}
\end{align*}
For both newly active and newly inactive site we may ignore the first term in the union since these correspond to rules that influence the output sites $\usite_j$ and $\usite_l$.
In Fig. M-2 (b) and (c) these rules would correspond to lines leading from input sites (top layer) to the newly active (blue) or newly inactive (white) sites.  
However, the outputs at these sites can be computed using Eq. (M-4) for newly active sites and simply set to 0 for newly inactive sites, and so we ignore them in updating the next layer. 
What remains are the contributions of the second term in the union which correspond to the magenta lines in the top layer of Fig. M-2 (b) and (c), which we define as 
$r_{\kidx,\text{act}}$ and $r_{\kidx,\text{inact}}$ respectively.

If we restrict the output sites $\usite$ to be sites that remain active, we may expand Eq. \eqref{eq:app:update} as:
\begin{align*}
\label{eq:app:update2}
    \tilde y_{1}^{t+1}(\usite,c)=& b_0(c)+\displaystyle\sum_{c'}\sum\limits_{\kidx\in\kset_0}\sum_{(\isite,\usite)\in R_{t+1,\kidx}}W_0(\kidx,c',c)\left(y_0^{t}(\isite,c')+\Delta(\isite,c')\right)\\
    =&b_0(c)+\displaystyle\sum_{c'}\sum\limits_{\kidx\in\kset_0}\sum_{(\isite,\usite)\in R_{t,\kidx}}W_0(\kidx,c',c)\left(y_0^{t}(\isite,c')+\Delta(\isite,c')\right)\\
    &-\displaystyle\sum_{c'}\sum\limits_{\kidx\in\kset_0}\sum_{(\isite,\usite)\in r_{\kidx,\text{inact}}}W_0(\kidx,c',c)\underbrace{\left(y_0^{t}(\isite,c')+\Delta(\isite,c')\right)}_{=0 \text{ for } \isite=\usite_l}\\
    &+\displaystyle\sum_{c'}\sum\limits_{\kidx\in\kset_0}\sum_{(\isite,\usite)\in r_{\kidx,\text{act}}}W_0(\kidx,c',c)(\underbrace{y_0^{t}(\isite,c')}_{=0 \text{ for } \isite=\usite_j}+\Delta(\isite,c'))
\end{align*}
Where we have used the fact that at newly inactive sites $y_0^{t}(\isite,c')+\Delta(\isite,c')=0$ and at newly active sites $y_0^{t}(\isite,c')=0$.
This can be simplified as:
\begin{align*}
    \tilde y_{1}^{t+1}(\usite,c)=& b_0(c)+\displaystyle\sum_{c'}\sum\limits_{\kidx\in\kset_0}\sum_{(\isite,\usite)\in R_{t,\kidx}}W_0(\kidx,c',c)\left(y_0^{t}(\isite,c')+\Delta(\isite,c')\right)\\
    &+\displaystyle\sum_{c'}\sum\limits_{\kidx\in\kset_0}\sum_{(\isite,\usite)\in r_{\kidx,\text{act}}}W_0(\kidx,c',c)\Delta(\isite,c')\\
    =&\underbrace{b_0(c)+\displaystyle\sum_{c'}\sum\limits_{\kidx\in\kset_0}\sum_{(\isite,\usite)\in R_{t,\kidx}}W_0(\kidx,c',c)y_0^{t}(\isite,c')}_{y_1^t(\usite,c)}\\
    &+\displaystyle\sum_{c'}\sum\limits_{\kidx\in\kset_0}\sum_{(\isite,\usite)\in R_{t,\kidx}}W_0(\kidx,c',c)\Delta_0(\isite,c')\\
    &+\displaystyle\sum_{c'}\sum\limits_{\kidx\in\kset_0}\sum_{(\isite,\usite)\in r_{\kidx,\text{act}}}W_0(\kidx,c',c)\Delta_0(\isite,c')\\
    =&y_1^t(\usite,c)+\displaystyle\sum_{c'}\sum\limits_{\kidx\in\kset_0}\sum_{(\isite,\usite)\in R_{t,\kidx}\cup r_{\kidx,\text{act}}}W_0(\kidx,c',c)\Delta_0(\isite,c')
\end{align*}
It remains to find the rules $(\isite,\usite)$ in $R_{t,\kidx}\cup r_{\kidx,\text{act}}$ which have a non-zero contribution to the sum, \emph{i.e.} for which $\Delta_0(\isite,c')$ is non-zero. The increment is exactly non-zero for $\isite\in\fset_0$, corresponding to the input site affected by the new event. Note that this site could either \emph{(i)} remain active (input for rule in $R_{t,\kidx}$), \emph{(ii)} become inactive (input for rule in $R_{t,\kidx}$) or \emph{(iii)} become active (input for rule in $r_{\kidx,\text{act}}$)). Therefore, the rules that have a non-zero contribution are the ones drawn as magenta lines in the top row of Fig. M-2 (a), (b) and (c) respectively, where we ignore rules with newly active or inactive sites output sites. These rules also correspond exactly to $\mathcal{R}_{\kidx,1}$ defined in Eq. M-7. The resulting update equation becomes:
\begin{align*}
    \tilde y_{1}^{t+1}(\usite,c)=&y_1^t(\usite,c)+\displaystyle\sum_{c'}\sum\limits_{\kidx\in\kset_0}\sum_{(\isite,\usite)\in\mathcal{R}_{\kidx,1}}W_0(\kidx,c',c)\Delta_0(\isite,c')
\end{align*}
This is exactly Eq. (M-9). By applying the non-linearity we arrive at Eq. (M-10).

Now let us consider how the update propagates to the next layer. For this we need to find $\fset_1$, \emph{i.e.} the input sites of layer 1 that change. These are exactly the updated output sites of layer 0.
Every $\isite\in\fset_0$ affects the output site $\usite$ for which $(\isite,\usite)\in \mathcal{R}_{\kidx,1}$. To be part of this rulebook $\usite=\isite-\kidx$ and so we see that $\isite-\kidx\in\fset_1$ for all $\kidx\in\kset_0$, which is exactly mirrored by Eq. (M-7).

To propagate updates through layer 1 we thus repeat the steps up until now, but only consider changes at sites $\fset_1$ instead of $\fset_0$. By iterating this procedure, all layers of the network can be updated. 
This concludes the proof.

\subsection{Representations and FLOP computation}
\label{sec:app:representations}
\subsubsection{Representations} As the proposed asynchronous framework does not require any specific input representation, we evaluate two event embeddings, which are sparse in time and space.
The two event representations tested for both tasks are the event histogram \cite{Maqueda18cvpr} and the event queue \cite{Tulyakov19iccv}.
The former creates a two-channel image by binning events with positive polarity to the first channel and events with negative polarity to the second channel.
This histogram is created for a sliding window containing a constant number of events.
Therefore, if an event enters or leaves the sliding window, an update site is created and propagated through the network.
The second representation called event queue \cite{Tulyakov19iccv} is applied in a sliding window fashion as well.
%
%
The event queue stores the timestamps and polarities of the incoming events in a queue at the corresponding image locations.
The queues have a fixed length of 15 entries and are initialised with zeros.
If a queue is full, the oldest event is discarded.
The four dimensional tensor containing the timestamps and polarities of up to 15 events is reshaped to a three dimensional tensor by concatenating the timestamps and polarities along the 15 entries.
The resulting three dimensional tensor has two spatial dimensions and a channel dimension with 30 entries.
\subsubsection{FLOP computation}
Tab. \ref{tab:app:flop} shows the number of FLOPs to perform different operation in the network for standard networks and our method. 
\begin{table}
\centering
\resizebox{0.9\textwidth}{!}{
\renewcommand{\arraystretch}{1.5}
\begin{tabular}{m{2.5cm}C{4.5cm}>{\centering\arraybackslash}m{4.5cm}}
   & Dense Layer & Sparse Layer\\
 \hline
 Convolution & 
 $H_{\text{out}} W_{\text{out}}  c_{\text{out}} (2  k^2 c_{\text{in}} -1)$ & 
 $N_{r} c_{\text{in}}(2 c_{\text{out}} + 1)$ \\
 \hline
 Max Pooling &
 $H_{\text{out}} W_{\text{out}} c_{\text{out}} k^2$ & 
 $N_{a} c_{\text{out}} k^2$\\
 \hline
 Fully Connected & $2 c_{\text{in}} c_{\text{out}}$ & $2 c_{\text{in}} c_{\text{out}}$\\
 \hline
 ReLU & $H_{\text{out}} W_{\text{out}} c_{\text{in}}$ & $N_{a} c_{\text{in}}$\\
\end{tabular}}
\caption{\label{tab:app:flop} FLOPs computation at each layer. Here $N_r$ are the number of rules at that layer and $N_a$ are the number of active sites.}
\end{table}
The FLOPs needed for a standard convolution is $H_{\text{out}} W_{\text{out}} c_{\text{out}}(2k^2c_{\text{in}}-1)$ excluding bias. This is the result of performing $k^2c_{\text{in}}$ multiplications and $k^2c_{\text{in}}-1$ additions for each pixel and each output channel resulting in the term found in the table. For our asynchronous sparse formulation we compute the number of operations by following Eqs. (M-8) and (M-9). Computing the differences in Eq. (M-8) results in $N_r c_{\text{in}}$ operations, where $N_r$ are the number of rules at that layer. The convolution itself uses $N_r c_{\text{in}}$ multiplications and $N_r (c_{\text{in}}-1)$ additions for each output channel, resulting in a total of $c_{\text{out}}N_r (2c_{\text{in}}-1)$ operations. Finally, from Eq. (M-9) additional $N_r c_{\text{out}}$ operations need to be expended to add these increments to the previous state. In total, this results in $N_{r} c_{\text{in}}(2 c_{\text{out}} + 1)$ operations.     
\subsection{Sensitivity on the Number of Events}
\label{sec:app:ablation}
Tab. \ref{tab:app:wind_size} shows the computational complexity in MFLOPS and test accuracy on N-Caltech101\cite{Orchard15fns} for both sparse and dense VGG13. The table shows that the test accuracy is maximized at 25'000 events for the sparse network and reaches a plateau for the dense network. At this number of events amount of computation in the sparse network is  $46\%$ lower than for the dense network. For this reason we selected 25'000 events for all our further experiments in the main manuscript. 
\begin{table}
\centering
\newcommand{\widtha}{1.5cm}
\resizebox{1\textwidth}{!}{
\begin{tabular}{m{2.2cm}C{\widtha}C{\widtha}C{\widtha}C{\widtha}C{\widtha}C{\widtha}C{\widtha}C{\widtha}C{\widtha}>{\centering\arraybackslash}m{\widtha}}
 &\multicolumn{2}{c}{64} & \multicolumn{2}{c}{256} & \multicolumn{2}{c}{5000} & \multicolumn{2}{c}{25000} & \multicolumn{2}{c}{50000} \\
  \cmidrule(lr){2-3} \cmidrule(lr){4-5} \cmidrule(lr){6-7} \cmidrule(lr){8-9} \cmidrule(lr){10-11}
  &  Accuracy & MFLOP & Accuracy & MFLOP &  Accuracy & MFLOP &  Accuracy & MFLOP &  Accuracy & MFLOP\\
 \hline
 Dense VGG13 & 0.257& 1621.2& 0.456& 1621.2 & 0.745& 1621.2& 0.761& 1621.2 & 0.766& 1621.2\\
 Sparse VGG13 & 0.247 & 224.2& 0.435& 381.0 & 0.734& 697.5& 0.745& 884.2 & 0.730& 959.2\\
\end{tabular}}
\caption{\label{tab:app:wind_size} Computational complexity and test accuracy on N-Caltech101\cite{Orchard15fns} for sparse and dense VGG13 and a varying number of events.}
\end{table}
\subsection{Qualitative Results on Object Detection}
\label{sec:qual_results_object_detection}
\begin{figure}
\centering
\def\colwidth{0.29\textwidth}
\newcolumntype{M}[1]{>{\centering\arraybackslash}m{#1}}
\begin{tabular}{m{1.0em} M{\colwidth} M{\colwidth} M{\colwidth} }
\begin{turn}{90}
\emph{N-Caltech101~\cite{Orchard15fns}}
\end{turn} & \includegraphics[width=\linewidth]{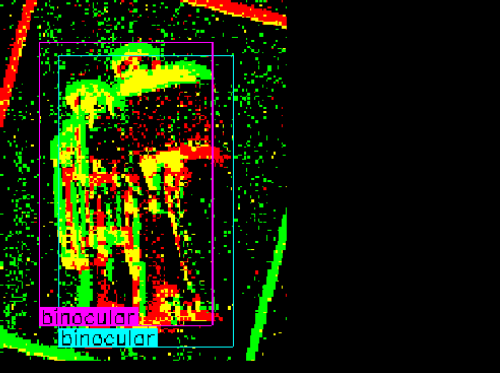}&
        \includegraphics[width=\linewidth]{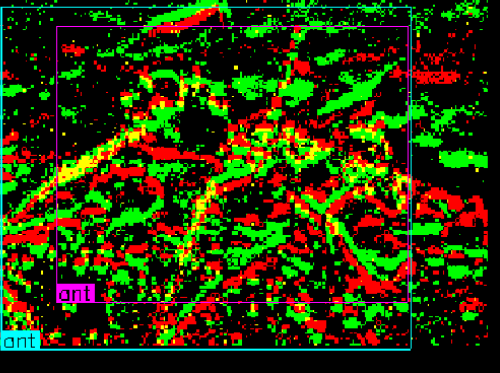}&
        \includegraphics[width=\linewidth]{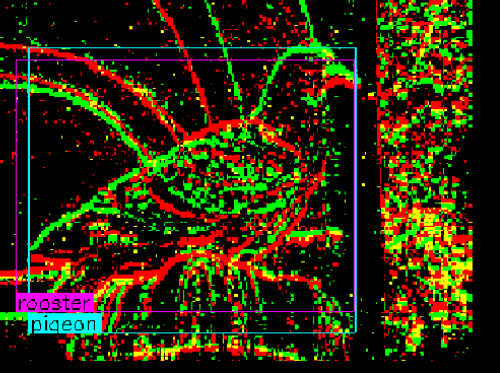}\tabularnewline
\begin{turn}{90}
\emph{Gen1 Auto.~\cite{tournemire2020large}}
\end{turn} &         \includegraphics[width=\linewidth]{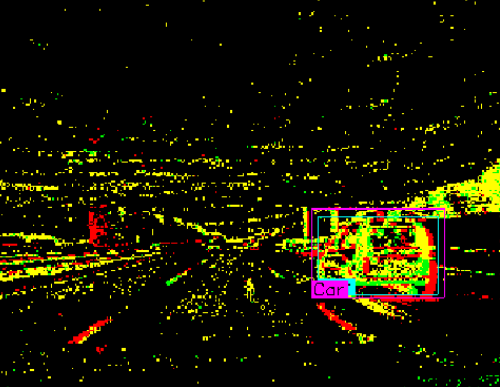}&
        \includegraphics[width=\linewidth]{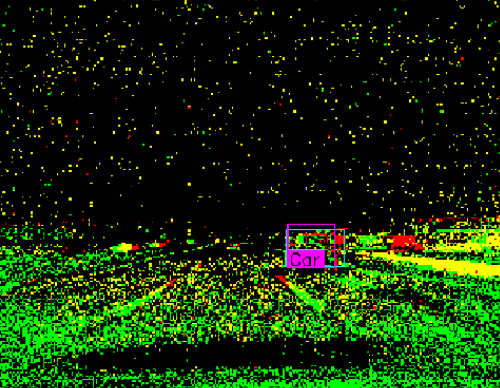}&
        \includegraphics[width=\linewidth]{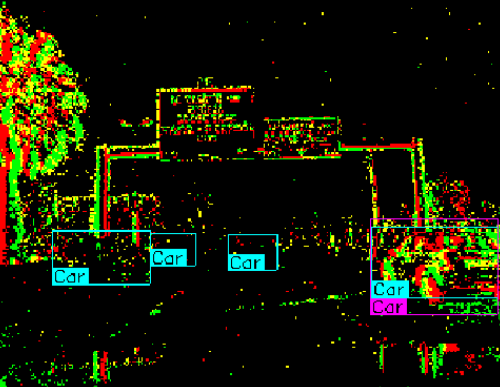} \tabularnewline
\end{tabular}
\addtolength{\tabcolsep}{4pt}
    \caption{Qualitative results of object detection (best viewed in color). Our predictions are shown in magenta, and labels in cyan. The first two columns present success cases, while the last column a failure case. On the first dataset, our method is mainly fooled by similar classes, such as "pigeon" and "rooster". In the second dataset, our approach detects cars generally well, but fails to detect the ones moving at similar speed due to the low event rate (bottom right). }
    \label{fig:experimental:det_qualitative}
\end{figure}

Here we show qualitative results of our method applied to the task of event-based object detection (Sec. M-4.2 and Tab. 4 in the main manuscript).
Failure cases of our approach include very similar classes, such as "rooster" and "pigeon" in the third column.
In the Gen1 Automotive dataset it can be seen that our approach works well for cars that are close and have a high relative motion. However, some cars are missed, especially if they have small relative speed and thereby only trigger few events (Fig.~\ref{fig:experimental:det_qualitative}, bottom right).
\begin{algorithm}
\caption{Asynchronous Sparse Convolution at layer n}
\label{alg:app:sparse_con1}
\begin{algorithmic}
\State \textbf{Update Active Sites}
\If{the first layer ($n=1$)}
\State - Update the active set $\aset_t$ with all new active and new inactive sites
\State - Initialize the rulebook $\mathcal{R}_{\kidx,0}=\emptyset$ and receptive field $\fset_0=\{\usite_i\}_i$.
\EndIf\\
\State \textbf{Update rulebook and receptive field}
\State - compute $\mathcal{R}_{\kidx,n}$ using Eq. (M-7) with $\mathcal{F}_{n-1}$
\State - compute $\fset_n$ using Eq. (M-6) with $\mathcal{F}_{n-1}$.\\
\State \textbf{Layer update}
\For {all $\usite$ in $\fset_n$}
\If {$\usite$ remains an active site}
\State - compute increment $\Delta_n$ using Eq. (M-8) with $y_{n-1}^t$ and $y_{n-1}^{t-1}$
\State - compute activation $\tilde y_{n}^t$ using Eq. (M-9) with $\Delta_n$ and $\tilde y_n^{t-1}$
\EndIf
\If {$\usite$ is newly active}
\State - compute activation $\tilde y_{n}^t$ using Eq. (M-4) with $y_{n-1}^t$ 
\EndIf
\If {$\usite$ is newly inactive}
\State - set activation $\tilde y_{n}^t$ to 0
\EndIf
\State - compute $y_n^t$ by applying a non-linearity as in Eq. (M-10)
\EndFor
\end{algorithmic}
\end{algorithm}

\end{document}